\documentclass[times,twocolumn,final,authoryear]{elsarticle}

\usepackage{ycviu}
\usepackage{framed,multirow}
\pdfoutput=1
\usepackage{amssymb}
\usepackage{latexsym}
\usepackage{graphicx}
\usepackage{amsmath}
\usepackage{amssymb}
\usepackage{booktabs}
\usepackage{multirow}
\usepackage{amsthm}
\usepackage{algorithm}  
\usepackage{algorithmic}  
\usepackage{CJKutf8}
\usepackage[capitalize]{cleveref}
\crefname{section}{Sec.}{Secs.}
\Crefname{section}{Section}{Sections}
\Crefname{table}{Table}{Tables}
\crefname{table}{Tab.}{Tabs.}
\newtheorem{thm}{Theorem}
\newtheorem{lemma}{Lemma}
\newtheorem{defn}{Definition}
\usepackage{url}
\usepackage{xcolor}
\definecolor{newcolor}{rgb}{.8,.349,.1}

\journal{Computer Vision and Image Understanding}

\begin{document}

\thispagestyle{empty}

\ifpreprint
  \setcounter{page}{1}
\else
  \setcounter{page}{1}
\fi

\begin{frontmatter}

\title{AdvFAS: a robust face anti-spoofing framework against adversarial examples}

\author[1]{Jiawei \snm{Chen}} 
\author[2]{Xiao \snm{Yang}}
\author[1]{Heng \snm{Yin}}
\author[1]{Mingzhi \snm{Ma}}
\author[4]{Bihui \snm{Chen}}
\author[4]{Jianteng \snm{Peng}}
\author[4]{Yandong \snm{Guo}}
\author[5]{Zhaoxia \snm{Yin}}
\author[2,3]{Hang \snm{Su}}
\cortext[cor1]{Corresponding author}
\ead{suhangss@mail.tsinghua.edu.cn}
\address[1]{Anhui Provincial Key Laboratory of Multimodal Cognitive Computation, Anhui University, Hefei 230601, China.}
\address[2]{Department of Computer Science and Technology, Institute for AI, THBI Lab, Tsinghua University, Beijing 100084, China.}
\address[3]{Zhongguancun Laboratory, Beijing, 100080, China}
\address[4]{OPPO Research Institute, Beijing, China}
\address[5]{Shanghai Key Laboratory of Multidimensional Information Processing, East China Normal University, Shanghai 200241, China.}

\received{1 May 2013}
\finalform{10 May 2013}
\accepted{13 May 2013}
\availableonline{15 May 2013}
\communicated{S. Sarkar}

\begin{abstract}
Ensuring the reliability of face recognition systems against presentation attacks necessitates the deployment of face anti-spoofing techniques. Despite considerable advancements in this domain, the ability of even the most state-of-the-art methods to defend against adversarial examples remains elusive. While several adversarial defense strategies have been proposed, they typically suffer from constrained practicability due to inevitable trade-offs between universality, effectiveness, and efficiency.
To overcome these challenges, we thoroughly delve into the coupled relationship between adversarial detection and face anti-spoofing. Based on this, we propose a robust face anti-spoofing framework, namely AdvFAS, that leverages two coupled scores to accurately distinguish between correctly detected and wrongly detected face images. Extensive experiments demonstrate the effectiveness of our framework in a variety of settings, including different attacks, datasets, and backbones, meanwhile enjoying high accuracy on clean examples. Moreover, we successfully apply the proposed method to detect real-world adversarial examples.

\end{abstract}


\end{frontmatter}

%\linenumbers

\section{Introdution}
\label{sec1}
The proliferation of face recognition systems has underscored the paramount importance of face anti-spoofing technology\citep{yang2014learn,wang2022learning,yu2022deep,yu2021dual,yang2019face}. This technology aims to distinguish between real human faces and various presentation attacks (face substitutes), such as print, video replay,  3D mask attacks, $etc$. 
As highlighted by \citep{szegedy2013intriguing}, deep neural networks are vulnerable to adversarial examples, which incorporate concealed messages into images. Furthermore, face recognition systems \citep{yang2021generating,yin2021adv,yang2021towards} based on deep neural networks also have exhibited their vulnerability against adversarial examples.
In terms of face anti-spoofing, adversarial examples can fool face anti-spoofing systems by introducing adversarial perturbations to the face substitutes, resulting in the paralysis of face recognition systems. 
Therefore, the creation of more realistic spoofing faces while encompassing a wide range of attack types remains a pressing concern in the domain of presentation attacks.

Moreover, the study of adversarial attacks is crucial to maximize the effect of presentation attacks \footnote{we do not consider real-to-fake attacks because they pose an ill-suited threat to face recognition systems}, where the breadth of attacks is a pivotal consideration. Adversarial attacks can be superimposed on existing face substitutes, endowing more examples with the ability to deceive face anti-spoofing technology. 

Therefore, in security-sensitive applications, face anti-spoofing must exhibit enough robustness against both presentation attacks and adversarial attacks. 

However, few studies have been concerned about defending adversarial examples in face anti-spoofing. \citep{bousnina2021unraveling} propose to defend differential evolution-based adversarial attacks using eight data augmentation techniques, which only concern the specific attack.  UniFAD\citep{deb2021unified} is proposed
 in an attempt to merge digital attacks and presentation attacks under one framework. 
However, UniFAD requires a large number of images to be prepared in advance to train an additional network. 
\begin{figure}[t]
  \centering
   \includegraphics[width=0.9\linewidth]{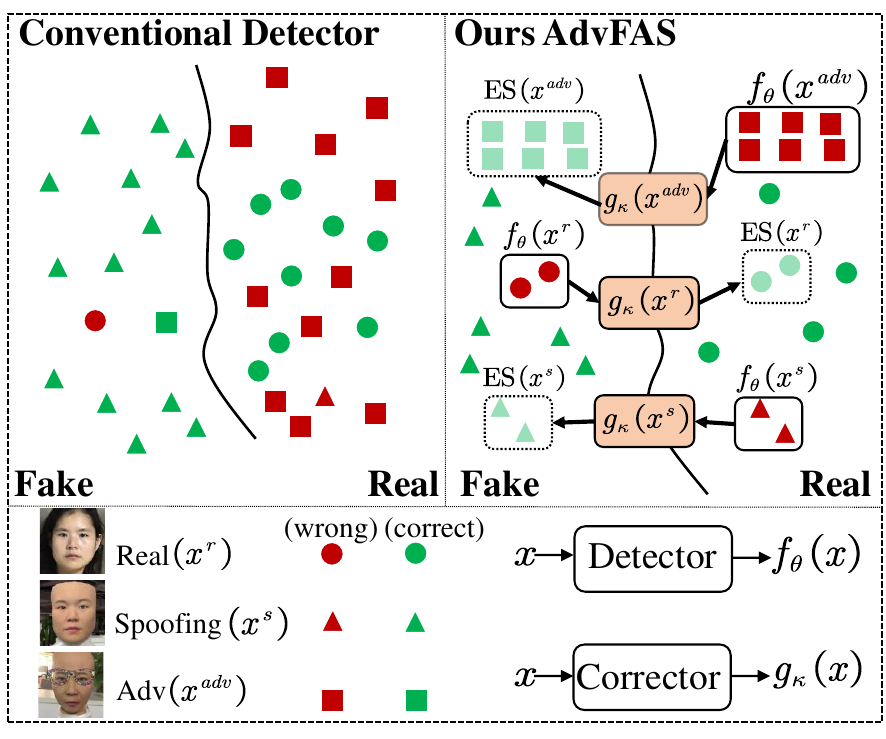}
   \caption{\textbf{Left:} Conventional detector methods demonstrate very poor performance in defending adversarial examples. \textbf{Right:} Our AdvFAS can distinguish wrongly detected examples from correctly detected ones through two coupled scores: $f_\theta(x)$ and ES$(x)$. ES($x$) refers to the expected score. $g_\kappa(x)$ builds the bridge between $f_\theta(x)$ and ES$(x)$.
}
   \label{fig: fig1}
\end{figure}
Another limitation is that both of them focus on detecting black-box attacks, which ignore the more challenging white-box attack detection. Compared to black-box attacks, it can only observe the output of the targeted model. An attacker has complete access to the target model in white-box settings, including the architecture and parameters of the model. Since black-box attacks are a more realistic assumption in the real world, evaluating models against white-box attacks is crucial to measure the performance of the model in the worst-case scenarios \citep{wiyatno1911adversarial}. 
Moreover, defenses against white-box attacks resist insiders, who could access the model by maliciously creating adversarial examples to jeopardize the performance of face recognition systems.

Adversarial training is recognized as an effective approach to defending against white-box attacks\citep{madry2017towards,zhang2019theoretically}. However, this approach typically leads to degraded detection performance on clean examples. 
This phenomenon is unacceptable in face anti-spoofing. 
A recent work\citep{pang2022two} proposes that adversarial examples could be provably told apart by two coupled rejection metrics. This approach performs excellently against white-box adversarial attacks and is compatible with different models for improving robustness. It inspired us to explore the coupled relationship between face anti-spoofing and adversarial detection. 

In this paper, we explore the coupled relationship between the two tasks, as illustrated in \cref{fig: fig1}, and propose a novel robust framework (AdvFAS) for face anti-spoofing to detect white-box adversarial examples. 
%\hangx{why detector and corrector. what is the benefit? A quick jump}
In \cref{sec: theoretical}, we present the theoretical analysis of this coupled relationship in detecting adversarial examples. 
Specifically, two coupled scores, $f_\theta(x)$ and ES$(x)$, are defined to explicitly represent the coupled relationship. 
$f_\theta(x)$ is the output score of the conventional detector, and ES$(x)$ is the expected score. 
In the case of correctly detected examples, ES$(x)$ is equal to $f_\theta(x)$. 
ES$(x)$ is equal to the ground truth label (0 or 1) for wrongly detected examples. However, due to the absence of the true label $y$, we are unable to obtain ES$(x)$ during testing. Therefore, a correction score $g_\kappa(x)$ is introduced to assist to predict ES$(x)$. We realize that the conventional detector structure is not sufficient to solve this problem with a single score $f_\theta(x)$, $i.e.$, an additional structure (named corrector) is required to calculate $g_\kappa(x)$. To save memory costs, we adopt a shared block with a two-head structure to model the detector and the corrector. The accuracy of clean examples is vulnerable to over-involvement of the adversarial examples during training. Thus, some strategies (such as stopping gradients and adding a mask) are adopted to eliminate the negative effects, which are illustrated in detail in \cref{sec: framework}.

To demonstrate the effectiveness of the framework, we conduct experiments on three widely used backbones (Depthnet \citep{depthnet}, Resnet-18 \citep{resnet}, and Resnet-50), two publicly available datasets (WMCA \citep{WMCA} and CASIA-SURF 3DMask \citep{casia3dmask}), and two adversarial attack methods (PGD \citep{madry2017towards} and patch attack \citep{tong2021facesec}).
In addition, we apply the proposed framework to CMFL \citep{rgbdmh} to verify its effectiveness on multi-modal inputs. 

Extensive experiments demonstrate that the proposed framework is suitable for various scenarios and obtains state-of-the-art performance. 
Moreover, adaptive attacks \citep{tramer2020adaptive} are designed to evade our corrector module, for the purpose of fully validating the effectiveness of the proposed framework. 
We further apply our framework to detect real-world adversarial examples, in order to show its practical applicability. %P6：实验结果

Our main contributions are summarized as follows:

\begin{itemize}
    \item ~We thoroughly explore the coupled relationship between face anti-spoofing and adversarial detection and propose a novel face anti-spoofing framework against adversarial examples. 

    \item ~We present a theoretical analysis that two coupled scores can distinguish wrongly detected face images from correctly detected ones.

    \item ~Extensive experiments show that the proposed framework is universal and effective in various scenarios. The highest accuracy on adversarial examples increased by 93.79\%.
\end{itemize}
\section{Related Work}
\textbf{Face Anti-Spoofing.}
There has been increasing research on face anti-spoofing in the last decade \citep{yu2022deep}.
During the first few years, traditional handcrafted features \citep{pan2007eyeblink,li2016generalized,freitas2012lbp,komulainen2013context,patel2016secure}  were utilized to detect print attacks. 
These include liveness cues, eye-blinking \citep{pan2007eyeblink,jee2006liveness,li2008eye}, head, face motion, and gaze tracking. 
However, these handcrafted features cannot be elegantly applied because they require rich prior knowledge and are vulnerable to being fooled by attacks with dynamic characteristics ($eg$, replay attacks).
With the development of deep learning, many researchers have turned to deep learning to achieve face anti-spoofing. Some research \citep{zhang2021face,li2021diffusing,ming2022vitranspad} focuses on using the whole RGB image to extract features. 
As illustrated by \citep{wang2022patchnet,grinchuk20213d}, the network sometimes loses local information when training with whole faces. 
These approaches crop the face images into patches, trying to capture more local information. 
As presentation attacks become more realistic, RGB images of real faces and face substitutes become more difficult to distinguish.
\citep{george2021cross,jia2021dual,wang2022conv} utilize multi-modal information ($eg$, deep and infrared) or other sensors. 
These methods impose additional demands on sensors or deep estimation methods. 
Some studies \citep{jia2020single,jia2021dual,liu2022causal} demonstrated that the generalizability of the face anti-spoofing models is poor. These studies mainly investigate generalization to unseen domains and unknown attack types.
In general, the research on face anti-spoofing tends to be diverse and practical. 
However, most face anti-spoofing research ignores the threat to face recognition systems from adversarial attacks.

\textbf{Adversarial Examples Defense.} The methods of defending against adversarial examples can be broadly categorized into two categories: adversarial training \citep{madry2017towards,zhang2019theoretically,wu2020adversarial} and adversarial detection \citep{metzen2017detecting,gong2017adversarial,carrara2018adversarial,zhang2018detecting,ma2018characterizing,feinman2017detecting}. On the one hand, adversarial training is widely accepted as the most effective method to improve the adversarial robustness of deep learning models \citep{yu2022deep}. The main idea is to make the network robust by adding adversarial examples to the training process through a min-max formulation. \citep{madry2017towards} proposed the PGD-AT, which signiﬁcantly increased the adversarial robustness of deep learning models against a wide range of attacks. Based on PGD-AT, some derivative work has followed their designs and proposed settings such as AWP \citep{wu2020adversarial}, TRADES \citep{zhang2019theoretically}, and Free-AT \citep{shafahi2019adversarial}, $etc$. However, adversarial training generally reduces the detection accuracy on clean examples \citep{madry2017towards}, which is particularly evident in face anti-spoofing according to our experiments. On the other hand, adversarial detection bypasses the impact on the accuracy on clean examples by training an additional detector \citep{metzen2017detecting,gong2017adversarial,carrara2018adversarial,zhang2018detecting} or designing statistics \citep{ma2018characterizing,feinman2017detecting}. The pipeline of \citep{deb2021unified} is similar to adversarial detection. But training an additional network increases consumption, and the defense system will be destroyed when this network is attacked. 

Due to such open challenges, we propose a novel robust face anti-spoofing framework against adversarial examples (AdvFAS). Specifically, we find a coupled relationship between adversarial detection and face anti-spoofing, and use this relationship to merge the two tasks into a single framework.

\section{Method}
\begin{figure}[t]
  \centering
   \includegraphics[width=0.99\linewidth]{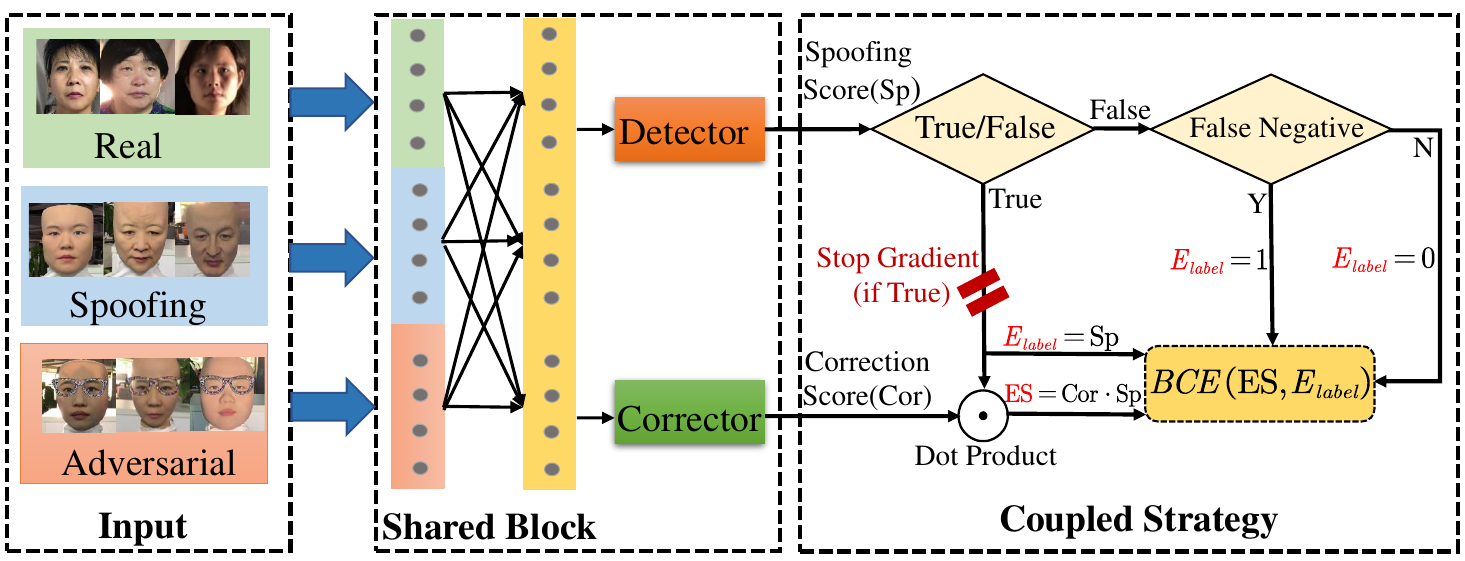}
   \caption{Construction of the objective $\mathcal{L}_{cor}$ in \cref{form: eight} for training
the corrector, which is the binary cross-entropy (BCE) loss between ES$(x)$ and $E_{label}$. The corrector shares a main backbone
with the detector, introducing little extra memory cost. 
}
   \label{fig: architecture}
\end{figure}

We propose a new robust framework for face anti-spoofing to detect adversarial examples by a corrector to correct the score of face anti-spoofing models, as illustrated in \cref{fig: architecture}.
Specifically, a two-head structure is adopted to implement the framework. One is a detector that does the conventional face anti-spoofing task; the other is a corrector that corrects the errors of the detector through a correction score.
The combination of the two can be used to calculate an expected score as the final score for judging whether a face is real.
In this section, we discuss the proposed framework in detail.
\subsection{Problem Formulation}

Defending against white-box adversarial attacks can always be formulated as a min-max problem and has also been verified to be effective. For a standard face anti-spoofing network (detector) $f_{\theta}$, consider a data pair $(x,y)$, with $x \in \mathbb{R}^d$ as the input and $y$ as the true label. Generally, the problem is formulated as a saddle point problem as follows:
\vspace{2ex}
\begin{equation}
        \setlength{\abovedisplayskip}{5pt}
        \setlength{\belowdisplayskip}{5pt}
    \underset{\theta}{\min}~\mathbb{E}_{p(x,y)}\left[\underset{\xi \in \mathcal{E}}{\max}\mathcal{L}(f_\theta(x+\xi),y)\right], 
        \label{form: first}
        \vspace{2ex}
\end{equation}
where the $x+\xi$ is the adversarial example generated by adding the perturbation $\xi$ from set $\mathcal{E} = \{\xi : \vert\vert\xi\vert\vert\leq\epsilon \}$ to the clean example $x$. In this paper, the coupled relationship of the two tasks can be similarly formulated as a min-max problem.

\textbf{What to Correct?} Some previous approaches for adversarial examples detection have associated adversarial characters with error detection. 
However, face anti-spoofing for clean examples still has a few errors despite impressive detection. 
And, for adversarially trained face anti-spoofing models, there are still many adversarial examples that can be 
correctly detected. Therefore,
in contrast to adversarial-based correction, it would be more reasonable to correct output based on whether the $f_{\theta}(x)$ will be wrong. 

\textbf{Couple Detector and Corrector.} The problem scenario is that the detector has excellent performance on clean examples with few errors, but has poor ability to detect adversarial examples. This inspires us not to wholly overturn current face anti-spoofing technologies but to correct some of the $f_\theta(x)$. Therefore, the corrector is created to tackle the problem and we construct the expected score (ES) as:%\hangx{a new concept of ESN. Why we need it and what problem you are solving.}
\vspace{2ex}
 \begin{equation}
 \vspace{2ex}
         \setlength{\abovedisplayskip}{5pt}
        \setlength{\belowdisplayskip}{5pt}
     {\rm ES} (x) = f_{\theta}(x)\cdot g_{\kappa}(x),
 \end{equation}
 where $g_{\kappa}(x) \in [0,1]$ is the corrector output 
 %\hangx{a new concept and why we need to define it} 
 and parameterized by ${\kappa}$. Through the ES, we expect the corrector to be aware of whether or not the $f_{\theta}(x)$ is correct and to correct the $f_{\theta}(x)$ when it is wrong. In other words, the ES, instead of $f_{\theta}(x)$, is treated as the final decision. In addition, the corrector and the detector are coupled: for a wrong $f_{\theta}(x)$, if the $f_{\theta}(x)$ is close to 
1, the ES$(x)$ must be close to 0 if the ES works (and vice versa). Because of this property, attackers cannot achieve successful attacks by attacking the detector or the corrector.
%\hangx{more details about ``coupled''} 

\textbf{Optimization Goal.} Assume that clean and adversarial examples are detected incorrectly as $x_c'$ and $x_a'$ respectively, where $x_a' =\underset{x'\in S(x)}{\arg\max}~\mathcal{L}_{spoof}(x',y;\theta)$. $S(x)$ is a set of allowed points around $x$~($e.g.$, a ball of $\vert\vert x'-x\vert\vert_p \le\epsilon$) and $x$ is spoofing. $\mathcal{L}_{spoof}$ means the loss of the detector.
Suppose $x^{*} \in \left\{ x_c',x_a' \right\}$ is the input that is detected wrongly by the detector%\hangx{what do you mean by ``detected by error?''}
. The problem is formulated as:
\vspace{2ex}
\begin{equation}
\vspace{2ex}
\label{eq: 3}
        \setlength{\abovedisplayskip}{5pt}
        \setlength{\belowdisplayskip}{5pt}
    \underset{\theta,\kappa}{\min}~\mathbb{E}_{p(x,y)}\left[\mathcal{L}_{spoof}(x,y;\theta)+\mathcal{L}_{cor}(x^*,y;\theta,\kappa)\right], 
\end{equation}
where the $\mathcal{L}_{cor}$ is utilized to optimize ES. Moreover, $f_{\theta}(x)$ and $g_{\kappa}(x)$ conform in distribution to the above coupled property by concurrently optimizing $\theta$ and $\kappa$. 
\cref{eq: 3} can be considered a min-max problem.
For the inner maximization step, the most aggressive perturbation is found for the face anti-spoofing network under the $\ell_p$ norm constraint to generate adversarial examples, $x_a'$. The outer minimization objective is to detect clean examples correctly and to correct the wrongly detected examples.        
More details are given in \cref{sec: theoretical} and \cref{sec: framework}.

\subsection{Theoretical Analysis}
\label{sec: theoretical}

The problem can be summarized into a set of cases $\mathcal{C}$.~$\mathcal{C}$ denotes all possible results for which the input $x$ is detected by the detector.~It can be split into 
$FP$: $\{x \vert x=spoofing||adv, x\in\mathbb{R}^d, f_{\theta}(x) \geq \frac{1}{2}\}$,
$FN$ : 
$\{x \vert x=real, x\in\mathbb{R}^d, f_{\theta}(x)<\frac{1}{2}\}$, 
$TN$ : $\{x \vert x=spoofing||adv, x\in\mathbb{R}^d, f_{\theta}(x)<\frac{1}{2}\}$, 
and~$TP$ : $\{x \vert x=real, x\in\mathbb{R}^d$, $f_{\theta}(x)\geq \frac{1}{2}\}$.~Where $adv$ are generated from spoofing faces. Clearly, $FP$ and $FN$ are the wrong cases, while $TN$ and $TP$ are the correct cases. 
 
At training time, we optimize ES by minimizing the binary cross-entropy (BCE) loss. Due to the diversity of cases and different optimization strategies of ES for different cases, an adaptive loss $\mathcal{L}_{cor}$ was designed:
\vspace{1ex}
  \begin{align}
        \setlength{\abovedisplayskip}{5pt}
        \setlength{\belowdisplayskip}{5pt}
    \label{form: 4}
     FP:  \mathcal{L}_{cor}(x;\theta,\kappa)&=BCE(f_{\theta}(x)\cdot g_{\kappa}(x),0),\\
          \label{form: 5}
     FN:  \mathcal{L}_{cor}(x;\theta,\kappa)&=BCE(f_{\theta}(x)\cdot g_{\kappa}(x),1),\\
     TN||TP:  \mathcal{L}_{cor}(x;\theta,\kappa)&=BCE(f_{\theta}(x)\cdot g_{\kappa}(x),f_{\theta}(x)).
     \label{form: 6}
 \end{align}
   \vspace{1ex}

Defining $E_{label}\in\{0,1,f_{\theta}(x)\}$, the adaptive loss can thereby be rewritten as:
  \vspace{2ex}
  \begin{equation}
        \setlength{\abovedisplayskip}{5pt}
        \setlength{\belowdisplayskip}{5pt}
     \mathcal{L}_{cor}(x;\theta,\kappa)= BCE(f_{\theta}(x)\cdot g_{\kappa}(x),E_{label}).
    \label{form: eight}
      \vspace{2ex}
 \end{equation}
 In this case, the optimal solution of minimizing $\mathcal{L}_{cor}$ is  $g_{\kappa}^*(x)=\frac{E_{label}}{f_{\theta}(x)}$. The corrector and the detector can be jointly optimized by minimizing:
   \vspace{2ex}
    \begin{equation}
    \label{eq: 8}
        \setlength{\abovedisplayskip}{5pt}
        \setlength{\belowdisplayskip}{5pt}
        \mathcal{L}_{cs} = \mathcal{L}_{spoof}(x;\theta) + \lambda\cdot \mathcal{L}_{cor}(x^*;\theta,\kappa).
          \vspace{2ex}
    \end{equation}
Here, $\lambda$ is a  hyper-parameter that defaults to 1.0. The \cref{eq: 8} has the same optimization objective as \cref{eq: 3}. In the training phase, the network simultaneously optimizes parameters for the detector and the corrector by $\mathcal{L}_{cs}$. The whole process is described in detail in Algorithm \ref{algorithm: one} .
\begin{algorithm}[t]
 \caption{the optimization strategy of AdvFAS}
  \hspace*{0.02in}{\bf Input:}{A detector $f_\theta$ with loss function $\mathcal{L}_{spoof}$ and a corrector $g_\kappa$ with a coupled loss $\mathcal{L}_{cor}$.}

   \hspace*{0.02in}{\bf Input:}{A clean example $x$ and ground-truth label $y$ (spoofing: 0, real: 1) and epochs $E$.}
  
  \hspace*{0.02in}{\bf Output:}{$f_\theta(x),~g_\kappa(x).$}
  
\begin{algorithmic}[1] 
\FOR{e = 0 to $E$-1}
\IF{y = 0}
\STATE Generate perturbation $\xi$ to $x$;
\STATE $x^* = x + \xi$;
\ENDIF
\STATE Put $x$ to the detector $f_\theta$;
\STATE Put $x$ and $x^*$ to the corrector $g_\kappa$;
\STATE Minimize the $\mathcal{L}_{spoof}$;
\IF{$f_\theta(x)$ or $f_\theta(x^*)$ is correct}
\STATE Stop the gradient of $f_\theta(x)$ or $f_\theta(x^*)$;
\ENDIF
\STATE Minimize the $\mathcal{L}_{cor}$.
\ENDFOR
\end{algorithmic}  
\label{algorithm: one}
\end{algorithm}

\subsubsection{Theoretically Coupling ES and $\boldsymbol{f_{\theta}}$}
\label{sec: 3.2.1}
Based on the above design of $\mathcal{L}_{cor}$, we state the optimization objectives of ES on different cases respectively.
We observe that properly coupling $f_{\theta}$ and ES can provably separate wrongly and correctly detected inputs, as stated below:  
\begin{lemma}
\label{lemma: 1}
    (Separability) Given the detector $f_{\theta}$,  $\forall x_1,x_2$ with spoofing scores larger than $\frac{1}{2}$, $i.e.$,
      \vspace{2ex}
    \begin{equation}
        f_\theta(x_1)\geq \frac{1}{2}, and~f_\theta(x_2)\geq \frac{1}{2}.
              \vspace{2ex}
    \end{equation}
    If $x_1$ is detected correctly, while $x_2$ is detected wrongly,
    then $\rm{ES}$ $(x_1)$ $> \frac{1}{2} > $ $\rm{ES}$ $(x_2)$.
\end{lemma}

\textit{Proof}. $x_1$ conforms to \cref{form: 6} because it is correctly detected, thus the ES$(x_1)$=$f_\theta(x_1)\geq \frac{1}{2}$. On the other side, $x_2$ is wrongly detected and $f_{\theta}(x_2)\geq \frac{1}{2}$, so it conforms to \cref{form: 4} and we have ES$(x_2)=0<\frac{1}{2}$. As for other cases, we have a similar conclusion.

\begin{lemma}
\label{lemma: 2}
    (Separability) Given the detector $f_{\theta}$,  $\forall x_1,x_2$ with spoofing scores less than $\frac{1}{2}$, $ie$,
      \vspace{2ex}
    \begin{equation}
        f_\theta(x_1) < \frac{1}{2}, and~f_\theta(x_2) < \frac{1}{2}.
          \vspace{2ex}
    \end{equation}
    If $x_1$ is detected correctly, while $x_2$ is detected wrongly,
    then $\rm{ES}$ $(x_1)$ $< \frac{1}{2} < $ $\rm{ES}$ $(x_2)$.
\end{lemma}

\textit{Proof}. $x_1$ conforms to \cref{form: 6} because it is correctly detected, thus the ES$(x_1)$=$f_\theta(x_1) < \frac{1}{2}$. On the other side, $x_2$ is wrongly detected and $f_{\theta}(x_2) < \frac{1}{2}$, so it conforms to \cref{form: 5} and we have ES$(x_2)=1 \geq \frac{1}{2}$. 

According to \cref{lemma: 1} (\cref{lemma: 2}), when the spoofing scores of the detector are larger (less) than $\frac{1}{2}$, then for any $x$ that passes the detector, ES$(x) \geq \frac{1}{2}$ (ES$(x) < \frac{1}{2}$) as long as $x$ is correctly detected; conversely, ES$(x) <\frac{1}{2}$ (ES$(x) \geq \frac{1}{2}$). The property of the two coupled scores can be utilized to distinguish wrongly detected examples from correctly detected ones. 

\subsubsection{Coupling in Practice ES and $\boldsymbol{f_{\theta}}$}
\cref{sec: 3.2.1} shows that coupling ES and $f_{\theta}$ is properly separable for wrongly and correctly detected inputs. However, during actual training, there is necessarily an error between $g_{\kappa}(x)$ and $g_{\kappa}^*(x)$. We introduce a definition on the point-wise error between $g_{\kappa}(x)$ and $g_{\kappa}^*(x)$:
\begin{defn}
\label{defn: 1}
    (Point-wisely $\delta$-error). If a point x satisfies the following range:
          \vspace{2ex}
    \begin{equation}
        |g_\kappa(x)- g_\kappa^*(x)| \leq \frac{\delta}{2},
        \label{form:10}
              \vspace{2ex}
    \end{equation}
    where $\delta \in [0,1)$, then $g_{\kappa}$ is called $\delta$-error at the input $x$. 
\end{defn}

For any $g_{\kappa}(x)$ trained to be better than a random guess at $x$,  the $\delta$ is always found to satisfy Definition \ref{defn: 1}. In particular, assume that $g_\kappa(x)$ performs a random guess on $x$, meaning $g_\kappa(x)=\frac{1}{2}$. Because $g_\kappa^*(x) \in [0,1]$, there is $|\frac{1}{2}- g_\kappa^*(x)| \leq \frac{1}{2}$, which means that even a random guess $g_\kappa(x)$ can satisfy Definition \ref{defn: 1} with $\delta=1$.

We prove the separability of wrongly and correctly detected inputs at the optimal threshold equal to $\frac{1}{2}$ in \cref{lemma: 1}. 
However, in practice we cannot access the optimal threshold, which is not always equal to $\frac{1}{2}$ due to training error. To this end, we demonstrate that the detector with $\frac{1}{2-\delta}$ and the corrector with $\delta$-error $g_\kappa(x)$ can be coupled to achieve separability, similarly to the property shown in \cref{lemma: 1}. 
\begin{thm}
\label{theorem: 1}
  (Separability) Given the detector $f_{\theta}$,  for any $x_1,x_2$ with spoofing scores larger than $\frac{1}{2-\delta}$, $i.e.$,
        \vspace{2ex}
\begin{equation}
    f_\theta(x_1)>\frac{1}{2-\delta}, ~and~f_\theta(x_2)>\frac{1}{2-\delta},
 \label{form: 11}
       \vspace{2ex}
\end{equation}
where $\delta \in [0,1)$, $x_1$ and $x_2$ are detected correctly and wrongly respectively, and $g_\kappa$ is $\delta$-error at $x_1$, $x_2$, then there must be  $\rm{ES}$ $(x_1) > \frac{1}{2} >$ $\rm{ES}$ $(x_2)$.
\end{thm}
Specifically, after we reset the threshold of the corrector to $\frac{1}{2-\delta}$ and obtain the remaining examples, any wrongly detected examples will be corrected correctly by the ES, as long as $g_\kappa$ is trained to be $\delta$-error for these examples.  

\textit{Proof}. The conditions in \cref{theorem: 1} can be written as $x_1$ is correctly detected, $f_\theta(x_1) > \frac{1}{2-\delta}$ and $x_2$ is wrongly detected, $f_\theta(x_2) > \frac{1}{2-\delta}$, where $\delta \in [0,1)$. Since $g_\kappa(x)$ is $\delta$-error at $x_1$ and $x_2$, according to \cref{defn: 1}, we have the following range:
      \vspace{2ex}
\begin{equation}
    |g_\kappa(x)- g_\kappa^*(x)| \leq \frac{\delta}{2}.
          \vspace{2ex}
\end{equation}
For $x_1$, there is $g_\kappa^*(x_1)=1$. And we can obtain
      \vspace{2ex}
\begin{align}
    {\rm ES}(x_1) &= f_\theta(x_1) \cdot g_\alpha(x_1)\nonumber
                    > f_\theta(x_1) \cdot (\frac{2-\delta}{2})\\
                    &> (\frac{1}{2-\delta}) \cdot (\frac{2-\delta}{2})=\frac{1}{2}.
                          \vspace{2ex}
\end{align}
Similarly for $x_2$, there is $g_\kappa^*(x_2)=0$. 
We can obtain:
      \vspace{2ex}
\begin{align}
    {\rm ES}(x_2) &= f_\theta(x_2) \cdot g_\alpha(x_2)\nonumber
    < f_\theta(x_2) \cdot \frac{\delta}{2}\\
    &<\frac{1}{2-\delta} \cdot \frac{\delta}{2}<\frac{1}{2}.
          \vspace{2ex}
\end{align}
Thus we have proven ES$(x_1) > \frac{1}{2} >$ ES$(x_2)$ and the \cref{theorem: 1} is established.
\subsection{Framework Details}
\label{sec: framework}
\textbf{Coupling Loss Design.} The $\mathcal{L}_{cs}$ consists of $\mathcal{L}_{spoof}$ and  $\mathcal{L}_{cor}$. The former is  the same as the conventional face anti-spoofing methods and the latter is utilized to optimize the corrector. Therefore, the conventional loss of face anti-spoofing is not altered during training. In other words, the model degenerates to the regular face anti-spoofing model when $\lambda$ is equal to 0. The particularity ensures that the proposed framework is compatible with different face anti-spoofing frameworks. (In the case of the loss function, it is sufficient to add $\mathcal{L}_{cor}$). 

\textbf{Shared Block.} There are three types of possible inputs: adversarial examples generated from spoofing faces, spoofing faces, and real faces. 
During the training period, all three types of images are shuffled and then input into the shared block.
The shared block simultaneously optimizes parameters for the detector and the corrector.
Compared to training on an additional network, the design of a shared block saves computation and memory costs.

\textbf{Stop Gradients and Add a Mask.} When the images are detected correctly, stopping gradients on the spoof scores of $\mathcal{L}_{cor}$ can avoid focusing on easy examples and keep the optimal solution of the detector unbiased. In particular, we use $p(x,y)$ to represent data distribution. For $\mathcal{L}_{spoof}$, the optimal solution of minimizing $\mathcal{L}_{spoof}(f_\theta(x),y)$ is $f_\theta(x)[y] = p(y|x)$. However, if we do not stop gradients of $f_\theta(x)[y]$ in the corrector, then the optimal $\theta$ of the entire framework objective no longer satisfies  $f_\theta(x)[y] = p(y|x)$. In this case, the corrector will introduce bias in the optimal solution of the detector. A mask is utilized to eliminate the loss of adversarial examples to ensure that the corrector works as conventionally as possible during training.
Specifically, the mask, which is an image-level $(0,1)$ matrix, is implemented to separate adversarial examples from clean examples (real and spoofing). 
The zeros in the matrix correspond to the adversarial examples. 
For a mask $\mathcal{M}=\{m_i\}_{i=1}^n$, where $n$ represents the number of images and $m_i\in \{0,1\}$ refers to the $i$-th image. Define $x_i$ as i-th examples. We can formulate as:
      \vspace{2ex}
\begin{equation}
        \setlength{\abovedisplayskip}{5pt}
        \setlength{\belowdisplayskip}{5pt}
\mathcal{L}_{spoof}(x_i,\theta) = m_i \cdot {L}_{spoof}(x_i,\theta). 
      \vspace{2ex}
\end{equation}
Without the mask, there may be problems caused by data imbalance. During the training time, too much participation of the adversarial examples will affect the detection of clean examples.

\section{Experiments}
   % \vspace{-2ex}
\begin{table}[b]
          \caption{The success rates (\%) of adversarial attacks against three models we study. The adversarial examples are crafted for HRFP, CMFL and DBMnet respectively using C\&W, PGD, patch, DeepFool and AutoAttack on WMCA and CASIA-SURF 3DMask.}
    \begin{center}
    \small %footnotesize
    \setlength{\tabcolsep}{3pt}
    \begin{tabular}{c|ccc|ccc}
        \hline
        \multirow{2}{*}{Method}& 
        \multicolumn{3}{c|}{WMCA}&\multicolumn{3}{c}{ CASIA-SURF 3DMask}\\
        \cline{2-7}
          & HRFP & CMFL & DBMnet & HRFP & CMFL & DBMnet \\

          \hline
        C\&W  &90.10&100.0&100.0&71.80&100.0&100.0 \\
        PGD &87.40&97.55&100.0&80.00&99.82&99.65\\
        Patch &81.21&86.86&100.0&73.60&98.10&100.0\\
        DeepFool &90.00&100.0&100.0&74.20&100.0&100.0 \\
        AutoAttack      &97.00&100.0&99.78&88.40&100.0&99.78 \\
        \hline

    \end{tabular}
    \end{center}
          	    \label{tab:tab1}
\end{table}
\begin{table*}[t]
  	   \caption{Performance of the baseline systems and the proposed framework against PGD on WMCA. The performance is tested on three models Depthnet, Resnet18, and Resnet50 respectively. Note that only color channels are used on WMCA. The ACC (w/o adv) (\%) represents the accuracy of models in detecting clean examples. The ACC (adv) (\%) is the accuracy of models in detecting adversarial examples. The ACC(avg) refers to the average test accuracy on clean and adversarial examples.}
        \resizebox{\linewidth}{!}{
	\begin{tabular}{c|ccc|ccc|ccc}
		\hline
		\multirow{2}{*}{Method}& 
		\multicolumn{3}{c|}{Depthnet}&\multicolumn{3}{c|}{ Resnet18} &\multicolumn{3}{c}{Resnet50}\cr\cline{2-10}
		  & ACC(w/o adv) & ACC(adv)& ACC(avg)&ACC(w/o adv) & ACC(adv)&ACC(avg)& ACC(w/o adv) & ACC(adv) &ACC(avg)\cr
		\cline{1-10}
		Clean &\textbf{91.10}&22.15&56.63&87.58&5.59&46.59&\textbf{89.90}&20.33&55.12\\
		PGD-AT(eps=16/255)  &25.43&99.87&62.65&22.25&\textbf{100.0}&61.13&25.07&94.69&59.88\\
		PGD-AWP(eps=16/255) &34.94&78.68&56.81&29.31&95.84&62.58&22.14&\textbf{100.0}&61.07\\
		PGD-AT(eps=8/255)   &22.29&\textbf{100.0}&61.15&27.4&90.23&58.82&20.99&99.94&60.47\\
	    PGD-AWP(eps=8/255)   &24.00&98.20&66.10&79.27&97.26&88.27&30.06&\textbf{100.0}&65.03\\
		Ours                &87.32&92.01&\textbf{\textcolor{red}{89.67}}&\textbf{90.26}&98.03&\textbf{\textcolor{red}{94.15}}&88.55&97.80&\textbf{\textcolor{red}{93.18}}\\
		\hline
	\end{tabular}
 }
    	  \label{tab:tab2}
\end{table*}
\begin{table}[t]
    	   \caption{Performance of the baseline systems and the proposed framework against PGD on WMCA. The performance is tested on CMFL.}
    \begin{center}
    \small %footnotesize
    \setlength{\tabcolsep}{3pt}
    \begin{tabular}{c|ccc}
        \hline
		\multirow{2}{*}{Method}& \multicolumn{3}{c}{CMFL}\\
		\cline{2-4}
		  & ACC(w/o adv) & ACC(adv) & ACC(avg) \\
		  \hline
		Clean  &\textbf{86.57}&2.11&44.34\\
		PGD-AT(eps=16/255) &21.83&99.89&60.86\\
      	PGD-AT(eps=8/255)  &79.27&\textbf{100.0}&89.64\\
        PGD-AWP(eps=16/255)  &23.50&\textbf{100.0}&61.75\\
        PGD-AWP(eps=8/255)  &24.89&99.81&62.35\\
		Ours  &86.10&95.90&\textbf{\textcolor{red}{91.00}}\\
		\hline

    \end{tabular}
    \end{center}
        	\label{tab:tab3}
\end{table}
\begin{table}[t]
    \caption{The AUC of the baseline systems and the proposed framework against PGD on CASIA-SURF 3DMask.}
    \vspace{-2ex}
    \begin{center}
        \scriptsize
        \setlength{\tabcolsep}{3pt}
        \resizebox{0.8\linewidth}{!}{
        \begin{tabular}{c|c|c}
            \cline{1-3}
            Model & Method & AUC(\%) \\
            \hline
            Depthnet & 
            \begin{tabular}{@{}c@{}}
                PGD-AT(eps=16/255) \\
                PGD-AT(eps=8/255) \\
                PGD-AWP(eps=16/255) \\
                PGD-AWP(eps=8/255) \\
                Our
            \end{tabular} & 
            \begin{tabular}{@{}c@{}}
                73.21 \\
                78.34 \\
                68.00 \\
                77.22 \\
                \textbf{\textcolor{red}{99.12}}
            \end{tabular} \\
            \hline
            CMFL & 
            \begin{tabular}{@{}c@{}}
                PGD-AT(eps=16/255) \\
                PGD-AT(eps=8/255) \\
                PGD-AWP(eps=16/255) \\
                PGD-AWP(eps=8/255) \\
                Our
            \end{tabular} &
            \begin{tabular}{@{}c@{}}
                75.19 \\
                80.63 \\
                72.71 \\
                75.50 \\
                \textbf{\textcolor{red}{90.94}}
            \end{tabular} \\
            \hline
        \end{tabular}
        }
    \end{center}
    \label{tab:tab4}
\end{table} 
\begin{table}[t]
    	   \caption{The performance of the baseline systems and the proposed framework aganist other adversarial attacks (MI-FGSM and DI-FGSM) on CASIA-SURF 3DMask. And the performance is tested on Depthnet.}
\begin{center}
	\small

	   \setlength{\tabcolsep}{3pt}
	\begin{tabular}{c|ccc}
		\hline
		\multirow{2}{*}{Method}& 
		\multicolumn{3}{c}{Depthnet}\\
		\cline{2-4}
		  & ACC(w/o adv) & ACC(adv)&ACC(avg) \\
		\hline
		
		MIFGSM-AT  &68.22&\textbf{99.24}&83.73\\
		Ours &\textbf{85.15}&93.59&\textbf{\textcolor{red}{89.37}}\\
		\hline
  	DIFGSM-AT  &71.18&\textbf{98.74}&84.96\\
		Ours &\textbf{94.39}&87.64&\textbf{\textcolor{red}{91.02}}\\
        \hline
	\end{tabular}
	\end{center}
  	  	\label{tab:another}
\end{table} 
\vspace{2ex}
\begin{table}[t]
    	   \caption{Performance of the baseline systems and the proposed framework against PGD on WMCA. The performance is tested on FAS-wrapper and Patchnet.}
\begin{center}
\resizebox{\columnwidth}{!}{
	\begin{tabular}{c|ccc|ccc}
		\hline
		\multirow{2}{*}{Method}& 
		\multicolumn{3}{c}{FAS-wrapper}&\multicolumn{3}{c}{Patchnet}\\
		\cline{2-7}
		  & ACC(w/o adv) & ACC(adv)&ACC(avg)& ACC(w/o adv) & ACC(adv)&ACC(avg) \\
		\hline
		Clean&\textbf{92.31}&2.36& 47.34&\textbf{91.64}&4.54&48.09\\
  		PGD-AT(eps=16/255)  &52.51&96.18& 74.35&53.68&98.39&76.04\\
		PGD-AT(eps=8/255)  &73.83&\textbf{99.04}& 86.44&74.16&97.46&85.81\\
    	PGD-AWP(eps=16/255)  &51.38&98.36& 74.87&58.39&\textbf{98.78}&78.59\\
  		PGD-AWP(eps=8/255)  &72.34&96.39& 84.37&73.81&98.56&86.19\\
		Ours &87.57&94.17&\textbf{\textcolor{red}{90.97}}&88.13&95.18&\textbf{\textcolor{red}{91.66}}\\
		\hline
	\end{tabular}
 }
	\end{center}
  	  	\label{tab:sota}
\end{table} 
% \vspace{2ex}
\subsection{Experimental Settings}
\textbf{Data Preparation.} The WMCA and CASIA-SURF 3DMask are used in our experiments as evaluation datasets. 
WMCA is a publicly available face image dataset that contains a variety of presentation attacks in 2D and 3D, such as paper print, video replay, and 3D mask.
We use the grandtest protocol of WMCA to divide it into the training set, validation set, and test set.
Compared with WMCA, face data on CASIA-SURF 3DMask is in video format.
Therefore, data pre-processing is required for the CASIA-SURF 3DMask. To obtain this, we extract one frame out of every three frames of each video and adopt MTCNN\citep{mtcnn} for face detection. 
Then, we crop and resize each extracted frame to the size of 128${\times}128$. 
Finally, we utilize an appropriate proportion to divide these pre-processing frames into the training set, the validation set, and the test set.      

\textbf{Network Setting.} The detector and corrector of the two-head structure adopt the same neural network, such as Depthnet, Resnet-18, and Resnet-50. 
Moreover, deep pixel-wise binary supervision \citep{george2019deep} is utilized to improve the performance of these neural networks for face anti-spoofing. 

\textbf{Attack Setting.} Five publicly available adversarial attacks are used in our experiments, including C\&W\citep{CW}, PGD, patch attack, DeepFool\citep{moosavi2016deepfool} and AutoAttack\citep{autoattack}.
C\&W, PGD, AutoAttack, and DeepFool are the most popular and powerful adversarial attacks in the digital world. In contrast, patch attack is easily realized in the physical world and has excellent attack capability.

\textbf{Training Setting.} 
The batch size is set to 100. 
The $\mathcal{L}_{cs}$ is optimized with the $\lambda$=1.0. 
An Adam optimizer with a learning rate (lr) of 1e-3 and a weight decay of 5e-5 is applied to our experiments.
We first base testing on a validation set to measure the optimal threshold. Then, the optimal threshold is utilized to calculate the final detection score on a test set.

\subsection{Vulnerability Evaluation of Existing Detectors}
\label{section: The Vulnerability of Existed SOTA Detectors}
To evaluate the vulnerability of existing SOTA face-spoofing methods, High Resolution Face Parts (HRFP)\citep{HRFP}, CMFL, and DBMnet \citep{dbmnet}.  are tested in adversarial attacks. HRFP achieved first place in the 3D High-Fidelity Mask Face Presentation Attack Detection Challenge based on the CASIA-SURF HiFiMask\citep{hifimask}. To ensure fairness, the checkpoint shared by the original authors is adopted in the evaluation.

As shown in \cref{tab:tab1}, we utilize adversarial attacks against these three methods mentioned above on WMCA and CASIA-SURF 3DMask respectively. 
Any success rate of adversarial attacks can simply be above 70\%, which shows that existing face-spoofing methods are vulnerable to adversarial attacks.
Especially under the attack configuration of C\&W, DeepFool, PGD and AutoAttack in CMFL and DBMnet, success rates of adversarial attacks can approach 100\% no matter on WMCA or CASIA-SURF 3DMask.
Although HRFP segments each face data into different face features as input to a corresponding convolution network, success rates of adversarial attacks also reach above 80\% on WMCA and 70\% on CASIA-SURF 3DMask.
Hence, the existing high-performance face-spoofing detectors can be easily broken by existing adversarial attack technologies, $i.e.$, it is necessary for face anti-spoofing methods to defend against adversarial examples.

\begin{table*}[htbp]
    	   \caption{Performance of the baseline systems and the proposed framework against patch attack on WMCA. The performance is tested on three models: Depthnet, Resnet18, and Resnet50. Note that only color channels are used on WMCA.}
        \resizebox{\linewidth}{!}{
	\begin{tabular}{c|ccc|ccc|ccc}
		\hline
		\multirow{2}{*}{Method}& 
		\multicolumn{3}{c|}{Depthnet}&\multicolumn{3}{c|}{ Resnet18} &\multicolumn{3}{c}{Resnet50}\\
	    \cline{2-10}
		  & ACC(w/o adv) & ACC(adv)& ACC(avg) & ACC(w/o adv) & ACC(adv)&ACC(avg)& ACC(w/o adv) & ACC(adv)&ACC(avg)\\
		\hline
		Clean
		   &91.10&3.90&47.93&87.58&10.26&51.07&\textbf{89.90}&28.91&58.74\\
		patch-AT(eps=255/255)
		    &21.04&99.99&60.52&21.80&\textbf{99.98}&60.89&20.94&99.93&60.44\\
		patch-AWP(eps=255/255)
		    &20.77&99.83&60.30&20.87&99.94&60.41&21.90&\textbf{100.0}&60.95\\
		patch-AT(eps=125/255)
		    &30.97&99.98&65.48&21.32&99.85&60.59&79.27&\textbf{100.0}&\textbf{\textcolor{red}{89.64}}\\
	    patch-AWP(eps=125/255)
		    &79.17&\textbf{100.0}&\textbf{\textcolor{red}{89.58}}&26.90&98.70&62.80&24.57&\textbf{100.0}&62.28\\
		Ours          &\textbf{92.16}&74.78&83.47&\textbf{93.31}&90.04&\textbf{\textcolor{red}{91.68}}&89.01&87.8&88.41\\
		\hline
	\end{tabular}
 }
    	       \label{tab:tab5}
\end{table*}
\begin{figure}[t]
  \centering
  \includegraphics[width=0.7\linewidth]{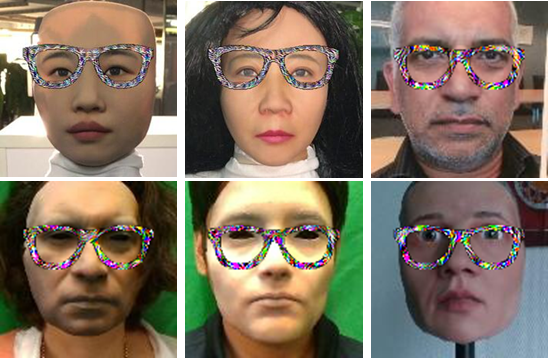}
  \caption{The adversarial examples are crafted on WMCA and CASIA-SURF 3DMask by patch attack.}
  \label{fig: patch}
\end{figure}
\subsection{Performance Against Different Settings}
\label{sec: different settings}
\textbf{AdvFAS~vs.~PGD.} The performance of the proposed framework and baselines against PGD in the grandtest protocol of WMCA are shown in \cref{tab:tab2}, \cref{tab:tab3} and \cref{tab:sota}.
In order to fully evaluate, we also use the AUC to evaluate our method and the baseline in \cref{tab:tab4}.  
We test baselines and AdvFAS on four models: depthnet, resnet18, resnet50 and CMFL. For CMFL, we use the four channels of RGBD on WMCA. 
To provide a comprehensive comparison, we also conduct a comparison with FAS-wrapper\citep{guo2022multi} and Patchnet\citep{wang2022patchnet}.
Due to their poor robustness, when training with clean examples, they have no ability to distinguish between adversarial examples and clean examples, although it has nice accuracy on the clean examples. The accuracy on adversarial examples is only around 20\%.
On the contrary, PGD-AT and PGD-AWP can recognize adversarial examples with higher accuracy, but these methods push the classification boundary of the training network away from the clean examples, so the accuracy on the clean examples dramatically reduces.
Our AdvFAS not only performs well on clean examples but also has higher accuracy on adversarial examples comparable to the above methods.

\textbf{AdvFAS~vs.~Patch Attack.} Patch attack is different from a slight noise perturbation adversarial attacks such as PGD, C\&W, DeepFool and AutoAttack. The adversarial examples perturbed by patch attack have a specific shape and are much easier to implement in the physical world. In \cref{fig: patch}, we show some demos of patch attacks. 
\begin{figure}[t]
        \center
        \scriptsize
        \begin{tabular}{cc}
                \includegraphics[width=3.1cm]{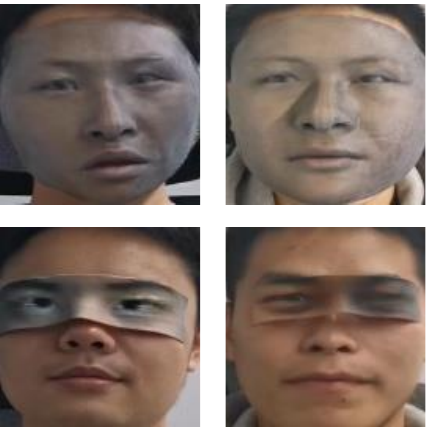} &    \includegraphics[width=3.1cm]{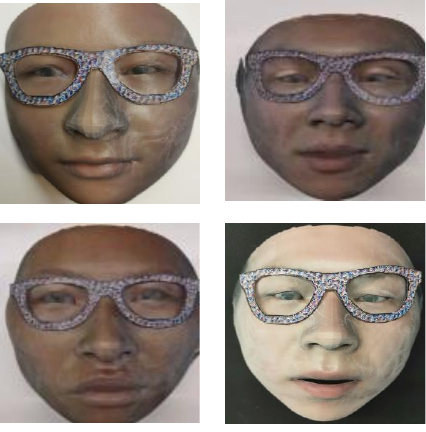}        \\
                (a) adversarial 3D face& (b) adversarial 2D glasses
        \end{tabular}
        \caption{Examples of test data in real-world scenarios; a) adversarial 3D face crafted for a face recognition system, b) adversarial 2D glasses crafted for a face anti-spoofing system.}
         \label{fig: 4}
\end{figure}

We conduct similar experiments in the configuration for patch attacks. The result is shown in \cref{tab:tab5} (We also try to test the methods on MI-FGSM and DI-FGSM adversarial attacks in \cref{tab:another}).
According to the experimental results, when training with the clean examples, although it has excellent accuracy on the clean examples, the accuracy on face images with patch attack is only about 14\% on average. Furthermore, different adversarial training strategies such as patch-AT and patch-AWP can recognize adversarial examples with excellent accuracy, but the accuracy of the clean examples dramatically reduce.
Our method significantly improves the metric for detecting adversarial examples while maintaining high performance on clean examples.

\begin{table}[t]
  \caption{Performance of the baseline systems and the proposed framework against PGD on CASIA-SURF 3DMask.}
\begin{center}
	\small

	   \setlength{\tabcolsep}{3pt}
	\begin{tabular}{c|ccc}
		\hline
		\multirow{2}{*}{Method}& 
		\multicolumn{3}{c}{Depthnet}\\
		\cline{2-4}
		  & ACC(w/o adv) & ACC(adv)&ACC(avg) \\
		\hline
		
		Clean  &99.59&31.30&65.45\\
		PGD-AT(eps=16/255) &31.88&99.71&65.80\\
      	PGD-AT(eps=8/255) &68.45&\textbf{100.0}&84.23\\
        PGD-AWP(eps=16/255) &31.40&99.46&65.43\\
        PGD-AWP(eps=8/255) &31.44&\textbf{100.0}&65.72\\
		Ours &\textbf{99.75}&99.88&\textbf{\textcolor{red}{99.82}}\\
		\hline
	\end{tabular}
	\end{center}
  	  	\label{tab:tab6}
\end{table} 

\textbf{On CASIA-SURF 3DMask}. To investigate the generality of the proposed framework, we compare with baselines on the second dataset, namely CASIA-SURF 3DMask. 
We adopt Depthnet as the basic detector, and design experiments to compare AdvFAS with the baseline method, as shown in \cref{tab:tab6}.
Both the ACC(w/o adv) and ACC(avg) of the proposed framework provide the best performance compared to the baseline methods. In fact, the ACC(adv) of our framework is close to 100\%. 

We attribute the excellent performance of AdvFAS to the coupled relationship between adversarial detection and face anti-spoofing and the training strategy. At training time, the coupled relationship, which can provably distinguish wrongly detected face images from correctly detected ones, can greatly improve the performance of detecting adversarial examples.
Our training strategy and the two-head structure bring a small modification to the conventional detector. Therefore, the performance of clean examples fluctuates in a small range. Moreover, because all wrongly detected examples are corrected (not just the adversarial examples), the accuracy of clean examples will also improve if the corrector is well optimized.
The experimental results of AdvFAS on different adversarial attacks, two publicly available datasets, and widely used backbones are similar.  
Therefore, our proposed method is efficient in improving the robustness of face anti-spoofing and is compatible with various face anti-spoofing methods.
\subsection{In Real-world Scenarios}
In this section, we apply our framework to detect adversarial examples in real-world scenarios, as illustrated in \cref{fig: 4}. We craft adversarial examples for a face recognition system ArcFace\citep{deng2019arcface} and a face anti-spoofing system, respectively, to validate the effectiveness of our framework against adversarial attacks in real-world scenarios. 

\textbf{For the Face Recognition System.} Adversarial 3D masks and adversarial 3D patches are crafted to fool face recognition systems. 
The adversarial 3D patches are crafted through \citep{yang2022controllable}, which simulates the complex transformations of faces in the physical world via 3D-face modeling. We employ some volunteers and they are required to wear the adversarial 3D face to record face videos for about 10 seconds. Then, we extract one frame out of every ten frames of each video. Finally, we adopt MTCNN to get 1555 face images as the input images. 

\textbf{For the Face Anti-Spoofing System.} We conduct a similar experimental configuration of \cref{sec: different settings} for patch attack. First, we use patch attacks to generate adversarial 2D glasses for the CMFL. Then, the adversarial examples are printed and cropped to the adversarial 2D glasses. Next, the adversarial 2D glasses are put on the 3D mask to record videos, extract frames, and crop to face images. Finally, the cropped face images are used as the input images of our proposed framework. 
\begin{table}[t]
	  \caption{Performance in real-world scenarios}
   \vspace{-2ex}
\begin{center}
       \resizebox{\linewidth}{!}{
	   \setlength{\tabcolsep}{3pt}
	\begin{tabular}{c|cccc}
		\hline
		\multirow{2}{*}{Method}& 
		\multicolumn{4}{c}{Depthnet}\\
		\cline{2-5}
		  & ACC(w/o adv) & ACC(adv 3D face) & ACC(adv 2D glasses) &ACC(avg)\\
		\hline
		
		Clean  &\textbf{85.96}&\textbf{93.21}&20.37&66.51\\
		Patch-AT(eps=16/255) &8.70&92.55&\textbf{89.34}&63.53\\
		Ours &85.08&78.09&88.48&\textbf{\textcolor{red}{83.58}}\\
		\hline
	\end{tabular}
	}
	\end{center}
		\label{tab:tab7}
\end{table} 

The experimental results are shown in \cref{tab:tab7}. Overall, our proposed framework outperforms the baseline methods. Although, due to the influence of lighting, position, and background, the proposed framework and baseline seem to have a performance gap compared to the results in the digital world. These experimental results show that the proposed framework still detects adversarial examples successfully in real-world scenarios. Furthermore, the results on the adversarial 3D face show that the face anti-spoofing systems have a certain defense against adversarial examples that can fool the face recognition systems.
\begin{figure}[t]
  \centering
  \includegraphics[width=0.8\linewidth]{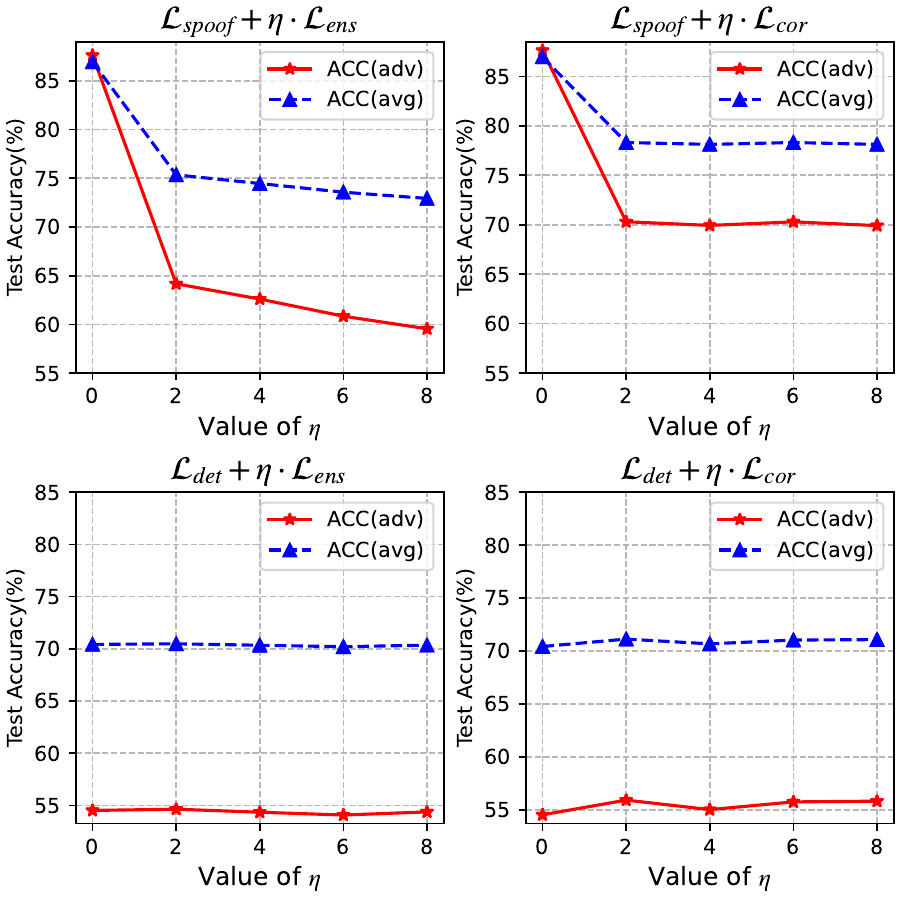}

  \caption{Performances under adaptive attacks on WMCA. We design four adaptive objectives to evade both the detector and corrector.}
  \label{fig: adaptive_attack}
\end{figure}
\subsection{Performance Against Adaptive Attacks}
We design adaptive attacks to simultaneously evade the detector and corrector. \cref{fig: adaptive_attack} shows the performance of the proposed framework under adaptive adversarial attacks on WMCA. We explore four different adaptive objectives, including $\mathcal{L}_{spoof}+\eta \cdot \mathcal{L}_{ens}$, $\mathcal{L}_{spoof}+\eta \cdot \mathcal{L}_{cor}$, $\mathcal{L}_{det}+\eta \cdot \mathcal{L}_{ens}$, and $\mathcal{L}_{det}+\eta\cdot\mathcal{L}_{cor}$, where $\mathcal{L}_{det}$ and $\mathcal{L}_{cor}$ is used to directly optimize the spoofing score and correction score respectively.
Test Accuracy in each sub-graph represents the detection accuracy, and the horizontal axis represents the value of the coefficient $\eta$ of $\mathcal{L}_{ens}$.
ACC(adv) is the test accuracy on adversarial examples crafted on WMCA, and ACC(avg) represents the average test accuracy on clean and adversarial examples.
When $\mathcal{L}_{spoof}$ and $\mathcal{L}_{ens}$, or $\mathcal{L}_{spoof}$ and $\mathcal{L}_{cor}$ are combined as the  adaptive attack objective, although ACC(adv) and ACC(avg) decrease at first, the latter trend tends to stabilize with the increase of $\eta$.  
When we combined $\mathcal{L}_{det}$ and $\mathcal{L}_{ens}$, or $\mathcal{L}_{det}$ and $\mathcal{L}_{cor}$, even though $\eta$ is increasing, the values of ACC(adv) and ACC(avg) are relatively stable.

Although there is some decrease in the ACC(adv) and ACC(avg), the ACC(avg) can still reach 70\% or more and maintain high performance in detecting clean examples, which means the proposed framework still significantly outperforms baselines. This can be attributed to the coupled mechanism. Specially, when adaptive attacks simultaneously evade the detector and corrector, the accuracy on clean and adversarial examples does not decrease significantly because of the coupled mechanism.
%coupled relationship 
\section{Conclusion}
In this paper, we first explore the coupled relationship between face anti-spoofing and adversarial detection and then propose a novel face anti-spoofing framework (AdvFAS) against adversarial examples. 
The effectiveness of the coupled relationship in detecting adversarial examples is proved from both theoretical and practical aspects.
By introducing two coupled scores to distinguish incorrectly detected face images from correctly detected ones, our framework significantly improves the robustness of face anti-spoofing technology. 
We conduct extensive experiments on three backbones and a multi-modal method, two publicly available datasets, and different adversarial attacks to validate the effectiveness our framework. 
Compared to results without our framework, our best result shows that the accuracy on adversarial examples increased by 93.79\%. 
Experimental results demonstrate that our framework is compatible with various face anti-spoofing methods and maintains excellent performance.
Finally, we detect real-world adversarial examples by the proposed framework and demonstrate its practical applicability. 

\section*{Acknowledgement}
This work was supported by the NSFC Projects (Nos. 62076147, 62071292, 62172001, 61771303, U19B2034, U1811461, U19A2081), Tsinghua Institute for Guo Qiang, Tsinghua-OPPO Joint Research Center for Future Terminal Technology, and the High Performance Computing Center, Tsinghua University.

\bibliographystyle{model2-names}
\bibliography{anti-spoofing}

\end{document}